\documentclass[11pt,a4paper]{article}

\usepackage[hyperref]{acl2019}
\usepackage{times}
\usepackage{latexsym}

\usepackage{comment}
\usepackage{amssymb}
\usepackage{todonotes}
\usepackage{url}
\usepackage{amsmath}
\usepackage{multirow}
\usepackage{graphicx}
\usepackage{epstopdf}
\usepackage{tabularx}
\usepackage{subcaption}
\usepackage{cleveref}
\usepackage{adjustbox}
\usepackage{float}
\usepackage{booktabs}
\usepackage{lingmacros}
\usepackage{xcolor}
\usepackage{makecell}
\usepackage{linguex}

\definecolor{cb-black}      {RGB}{  0,   0,   0}
\definecolor{cb-blue-green} {RGB}{  0,  073,  073}
\definecolor{cb-green-sea}  {RGB}{  0, 146, 146}
\definecolor{cb-rose}       {RGB}{255, 109, 182}
\definecolor{cb-salmon-pink}{RGB}{255, 182, 119}
\definecolor{cb-purple}     {RGB}{ 73,   0, 146}
\definecolor{cb-blue}       {RGB}{ 0, 109, 219}
\definecolor{cb-lilac}      {RGB}{182, 109, 255}
\definecolor{cb-blue-sky}   {RGB}{109, 182, 255}
\definecolor{cb-blue-light} {RGB}{182, 219, 255}
\definecolor{cb-burgundy}   {RGB}{146,   0,   0}
\definecolor{cb-brown}      {RGB}{146,  73,   0}
\definecolor{cb-clay}       {RGB}{219, 209,   0}
\definecolor{cb-green-lime} {RGB}{ 36, 255,  36}
\definecolor{cb-yellow}     {RGB}{255, 255, 109}

 \newcommand{\red}[1]{\textcolor{red}{#1}}

\aclfinalcopy 

\title{The Sensitivity of Language Models and Humans \\to Winograd Schema Perturbations}

\author{Mostafa Abdou\textsuperscript{$\dagger$}~~~Vinit Ravishankar\textsuperscript{$\clubsuit$} \AND ~~~Maria Barrett\textsuperscript{$\dagger$}~~~Yonatan Belinkov\textsuperscript{$\spadesuit$}~~~Desmond Elliott\textsuperscript{$\dagger$}~~~Anders S{\o}gaard\textsuperscript{$\dagger$} \\ 
  \textsuperscript{$\dagger$}Department of Computer Science, University of Copenhagen~~~\\ 
  \textsuperscript{$\clubsuit$}Department of Informatics, University of Oslo~~~ \\
  \textsuperscript{$\spadesuit$}John A. Paulson School of Engineering and Applied Sciences, Harvard University \\ \textsuperscript{$\spadesuit$}Computer Science and Artificial Intelligence Laboratory, Massachusetts Institute of Technology \\}

\date{}

\begin{document}
\maketitle
\begin{abstract}

Large-scale pretrained language models are the major driving force behind recent improvements in performance on the Winograd Schema Challenge, a widely employed test of commonsense reasoning ability. We show, however, with a new diagnostic dataset, that these models are sensitive to linguistic perturbations of the Winograd examples that minimally affect human understanding. Our results highlight interesting differences between humans and language models: language models are more sensitive to number or gender alternations and synonym replacements than humans, and humans are more stable and consistent in their predictions, maintain a much higher absolute performance, and perform better on non-associative instances than associative ones. Overall, humans are correct more often than out-of-the-box models, and the models are sometimes right for the wrong reasons. Finally, we show that fine-tuning on a large, task-specific dataset can offer a solution to these issues. 

\end{abstract}

\section{Introduction}
\label{intro}

Large-scale pre-trained language models have recently led to improvements across a range of natural language understanding (NLU) tasks \cite{devlin2018bert, radford2019language, yang2019xlnet}, but there is some scepticism that benchmark leaderboards do not represent the full picture \cite{kaushik2018much, jumelet2018language, poliak2018hypothesis}. An open question is whether these models generalize beyond their training data samples.  

In this paper, we examine how pre-trained language models generalize on the Winograd Schema Challenge (WSC). 

\begin{figure}
    \centering
    \includegraphics[width=0.5\textwidth]{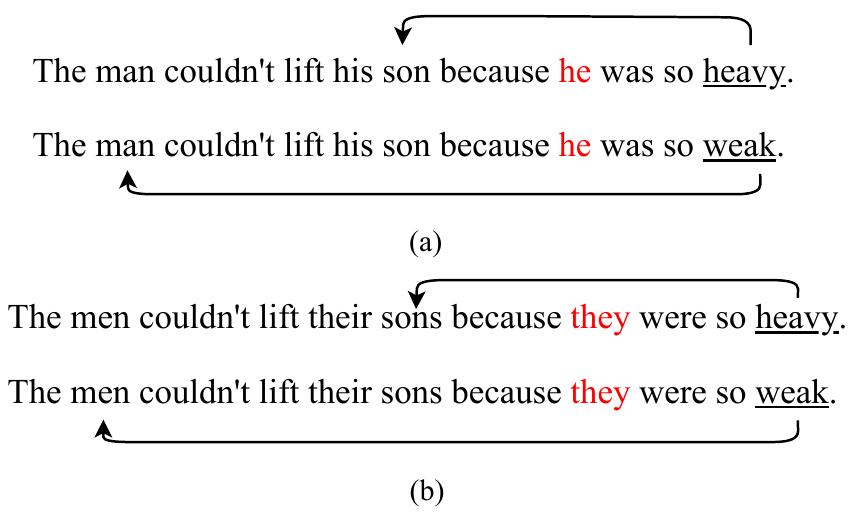}
    \caption{An example pair from the Winograd Schema Challange (a) and its perturbation (b). The \textcolor{red}{pronoun} resolves to one of the two referents, depending on the choice of the \underline{discriminatory segment}. The perturbation in (b) pluralizes the referents and the antecedents.}
    \label{fig:sentences}
\end{figure}

Named after Terry Winograd, the WSC, in its current form, was proposed by \newcite{levesque2012winograd} as an alternative to the Turing Test. The task takes the form of a binary reading comprehension test where a statement with two referents and a pronoun (or a possessive adjective) is given, and the correct antecedent of the pronoun must be chosen. Examples are chosen carefully to have a preferred reading, based on semantic plausibility rather than co-occurrence statistics. WSC examples come in pairs that are distinguished only by a \underline{discriminatory segment} that \textit{flips} the correct referent, as shown in Figure \hyperref[fig:sentences]{1a}. \citeauthor{levesque2012winograd} define a set of qualifying criteria for instances and the pitfalls to be avoided when constructing examples (see \S\ref{subsec:confounds}). These combine to ensure an instance functions as a test of what they refer to as `thinking' (or common sense reasoning).


Recent work has reported significant improvements on the WSC \cite{kocijan2019surprisingly, sakaguchi2019winogrande}. As with many other NLU tasks, this improvement is primarily due to large-scale language model pre-training, followed by fine-tuning for the target task.  We believe that further examination is warranted to determine whether these impressive results reflect a fundamental advance in reasoning ability, or whether our models have learned to simulate this ability in ways that do not generalize. 
In other words, do models
learn accidental correlations in our datasets, or do they extract patterns that generalize in robust ways beyond the dataset samples? 

In this paper, we conduct experiments to investigate this question. We define a set of lexical and syntactic variations and perturbations for the WSC examples and use altered examples (Figure \hyperref[fig:sentences]{1b}) to test models that have recently reported improved results. These variations and perturbations are designed to highlight the robustness of human linguistic and reasoning abilities and to test models under these conditions. 

\paragraph{Contributions} We introduce a new Winograd Schema dataset for evaluating generalization across seven controlled linguistic perturbations.\footnote{Code and dataset can be found at: \url{https://github.com/mhany90/enhanced_wsc/}} We use this dataset to compare human and language model sensitivity to those perturbations, finding marked differences in model performance. We present a detailed analysis of the behaviour of the language models and how they are affected by the perturbations. Finally, we investigate the effect of fine-tuning with large task-specific datasets, and present an error analysis for all models.

\section{Related Work}
\label{rel_work}
\paragraph{Probing datasets} Previous studies have explored the robustness of ML models towards different linguistic phenomena \citep{belinkov2019analysis}, e.g., by creating challenge datasets such as the one introduced here. When predicting subject-verb agreement, \citet{linzen2016assessing} found that inserting a relative clause hurt the performance of recurrent networks.\footnote{This contrasts with our results with Transformer-based architecture and is probably explained by memory loss in recurrent networks trained on short sequences. Similarly, \citet{gulordava2018colorless} tested whether a Recurrent Neural Network can predict long-distance number agreement
in various constructions comparing
natural and nonsensical sentences where RNNs
cannot rely on semantic or lexical cues.} 

A large body of research has since emerged on probing pre-trained (masked) language models for linguistic structure \cite{goldberg2019assessing,hewitt2019structural, lin2019open, clark2019does} and analysing them via comparison to psycholinguistic and brain imaging data \cite{abnar2019blackbox, ettinger2019bert, abdou2019higher, gauthier2019linking}. Other recent work has attempted to probe these models for what is referred to as \textit{common sense} or factual knowledge \cite{petroni2019language, feldman2019commonsense}. Their findings show that these models do indeed encode such knowledge and can be used for knowledge base completion or common sense mining from Wikipedia.

\paragraph{Clever Hans} A considerable amount of work has also been devoted to what might be described as the Clever Hans effect. This work has aimed to quantify the extent to which models are learning what we expect them to as opposed to leveraging statistical artifacts. This line of work has to date revealed significant problems (and some possible solutions to those problem) with reading comprehension datasets \cite{chen2016thorough, kaushik2018much}, natural language inference datasets \cite{tsuchiya2018performance, gururangan2018annotation, poliak2018hypothesis, belinkov2019adversarial, mccoy2019right}, and the story cloze challenge \cite{schwartz2017effect}, among others.

\paragraph{Winograd Schema Challenge}
\newcite{trinh2018simple} first proposed using neural language models for the WSC, achieving an accuracy of 63.7\% using an ensemble of 14 language models. \newcite{ruan2019exploring} and \newcite{kocijan2019surprisingly} fine-tune BERT \cite{devlin2018bert} on the PDP \cite{rahman2012resolving} and an automatically generated MaskedWiki dataset, reaching an accuracy of 71.1\% and 72.5\% respectively. Meanwhile, \newcite{radford2019language} report an accuracy of 70.7\% without fine-tuning using the GPT-2 language model. Most recently, \newcite{sakaguchi2019winogrande} present an adversarial filtering algorithm which they use for crowd-sourcing a large corpus of WSC-like examples. Fine-tuning RoBERTa \cite{liu2019roberta} on this, they achieve an accuracy of 90.1\%. 

In an orthogonal direction, \newcite{trichelair2018evaluation} presented a timely critical treatment of the WSC. They classified the dataset examples into associative and non-associative subsets, showing that the success of the LM ensemble of \newcite{trinh2018simple}  mainly resulted from improvements on the associative subset. Moreover, they suggested switching the candidate referents (where possible) to test whether systems make predictions by reasoning about the ``entirety of a schema'' or by exploiting ``statistical quirks of individual entities''. 

In a similar spirit, our work is a controlled study of robustness along different axes of linguistic variation. This type of study is rarely possible in NLP due to the large size of datasets used and the focus on obtaining improved results on said datasets. Like a carefully constructed dataset which is thought to require true natural language understanding, the WSC presents an ideal testbed for this investigation.

\section{Perturbations}
\label{perturbations}

\begin{table*}
\scriptsize
\centering
\renewcommand{\arraystretch}{1.3}
\begin{adjustbox}{width=\textwidth}
\begin{tabular}{p{2cm}p{9.5cm}l}
\toprule
 & Instance / Perturbed Instance                                                    & Count                                                                                                                                                                                                                   \\ \midrule
Original       & Sid explained his theory to Mark but he couldn't convince him. & 285                                                                      \\ \midrule
Tense       & Sid is explaining his theory to Mark but he can't convince him.                           & 281                                                                                                                        \\ \midrule
Number       & \textbf{Sid and Johnny} explained their theory to \textbf{Mark and Andrew} but they couldn't convince them.                           & 253                                                                                                                         \\ \midrule
Gender       &   \textbf{Lucy} explained her theory to \textbf{Emma} but she couldn't convince her.                        & 155                                                                                                                         \\ \midrule
Voice       & The theory was explained by Sid to Mark but he couldn't convince him.  & 220                                                                                                                      \\ \midrule
Relative clause  & Sid, \textbf{which we had seen on the discussion panel with Chris,} explained his theory to Mark but he couldn't convince him.  & 283 \\ \midrule

Adverb & Sid \textbf{diligently} explained his theory to Mark but he couldn't convince him.  & 283 \\ \midrule

Synonyms/Names & \textbf{John} explained his theory to \textbf{Jad} but he couldn't convince him. & 285  \\ 




\bottomrule
\end{tabular}
\end{adjustbox}
\caption{Examples from our dataset of the different perturbations applied to a WSC instance. }
\label{table:pert_examples}
\end{table*}

We define a suite of seven perturbations that can be applied to the 285 WSC examples, which we refer to as the original examples. These perturbations are designed to test the robustness of an answer to semantic, syntactic, and lexical variation. Each of the perturbations is applied to every example in the WSC (where possible), resulting in a dataset of 2330 examples, an example of each type is shown in Table \ref{table:pert_examples}. Crucially, the correct referent in each of the perturbed examples is \textbf{not} altered by the perturbation. The perturbations are manually constructed, except for the sampling of names and synonyms. Further details can be found in Appendix \ref{app:notes_perturbations}.

\paragraph{Tense switch (\textsc{ten})} 
Most WSC instances are written in the past tense and thus are changed to the present continuous tense (247 examples). The remaining 34 examples are changed from the present to the past tense. 

\paragraph{Number switch (\textsc{num})}
Referents have their numbers altered: singular referents (and the relevant pronouns) are pluralised (223 examples), and plural referents are modified to the singular (30 examples). Sentences with names have an extra name added via conjunction; eg. ``Carol'' is replaced with ``Carol and Susan''. Possessives only mark possession on the second conjunct (``John and Steve's uncle'' rather than ``John's and Steve's uncle''). 

\paragraph{Gender switch (\textsc{gen})}
Each of the referents in the sentence has their gender switched by replacing their names with other randomly drawn frequent English names of the opposite gender.\footnote{Names sourced from \url{https://github.com/AlessandroMinoccheri/human-names/tree/master/data}} 92\% of the generated data involved a gender switch for a name. Though humans may be biased towards gender \cite{collins2011content, desmond2010women, hoyle-etal-2019-unsupervised}, the perturbations do not introduce ambiguity concerning gender, only the entity.  101 examples were switched from male to female, and 55 examples the other way around. 

\paragraph{Voice switch (\textsc{vc})} 
All WSC examples, except for 210 and 211, are originally in the active voice and are therefore passivized. 210 and 211 are changed to the active voice. 65 examples could not be changed. Passive voice is known to be more difficult to process for humans \citep{olson1972comprehension, feng2015differences}.

\paragraph{Relative clause insertion (\textsc{rc})}
A relative clause is inserted after the first referent. For each example, an appropriate clause was constructed by first choosing a template such as ``who we had discussed'' or ``that is known for'' from a pre-selected set of 19 such templates. An appropriate ending, such as ``who we had discussed \underline{\textit{with the politicians}}'' is then appended to the template depending on the semantics of the particular instance. Relative clauses impose an increased demand on working memory capacity, thereby making processing more difficult for humans \cite{just1992capacity, gibson1998linguistic}. 

\paragraph{Adverbial qualification (\textsc{adv})}
An adverb is inserted to qualify the main verb of each instance. When a conjunction is present both verbs are modified. For instances with multiple sentences, all main verbs are modified. 

\paragraph{Synonym/Name substitution (\textsc{syn/na})}
Each of the two referents in an example is substituted with an appropriate synonym, or if it is a name, is replaced with a random name of the same gender from the same list of names used for the gender perturbation.

\subsection{Human Judgments}
\label{sec:data_collection}

We expect that humans are robust to these perturbations because they represent naturally occurring phenomena in language; we test this hypothesis by collecting human judgements for the perturbed examples. We collect the judgments for the perturbed examples using Amazon Mechanical Turk. 
The annotators are presented with each instance where the pronoun of interest is boldfaced and in red font. They are also presented with two options, one for each of the possible referents. They are then instructed to choose the most likely option, in exchange for \$0.12. Following \newcite{sakaguchi2019winogrande}, each instance is annotated by three annotators and majority vote results are reported. 
Results are reported later in \S\ref{analysis}. All three annotators agreed on the most likely option in 82-83\% of the instances, except for gender, where a full agreement was obtained for only 78\% of the instances. See Appendix \ref{app:humanresults} for further annotation statistics, a sample of the template presented to annotators, and restrictions applied to pool of annotators. We did not require an initial qualification task to select participants. 

\subsection{Confounds and Pitfalls}
\label{subsec:confounds}

Constructing WSC problems is known to be difficult. Indeed, the original dataset was carefully crafted by domain experts and subsequent attempts at creating WSC-like datasets by non-experts such as in \newcite{rahman2012resolving} have produced examples which were found to be less challenging than the original dataset. Two likely pitfalls listed in \newcite{levesque2012winograd} concern \textbf{A}) statistical preferences which make one answer more readily associated with the special discriminatory segment or other components of an example\footnote{\newcite{trichelair2018evaluation} find that 13.5\% of examples from the original WSC might still be considered to be \textit{associative}.}  (this is termed as \textit{Associativity}, and it is described as \textit{non-Google-proofness} in \newcite{levesque2012winograd}); and \textbf{B}) inherent ambiguity which makes the examples open to other plausible interpretations. In what follows, we discuss these pitfalls, demonstrating that the perturbed examples remain resilient to both.  

\begin{figure}[t]
    \centering
    \includegraphics[width=\columnwidth]{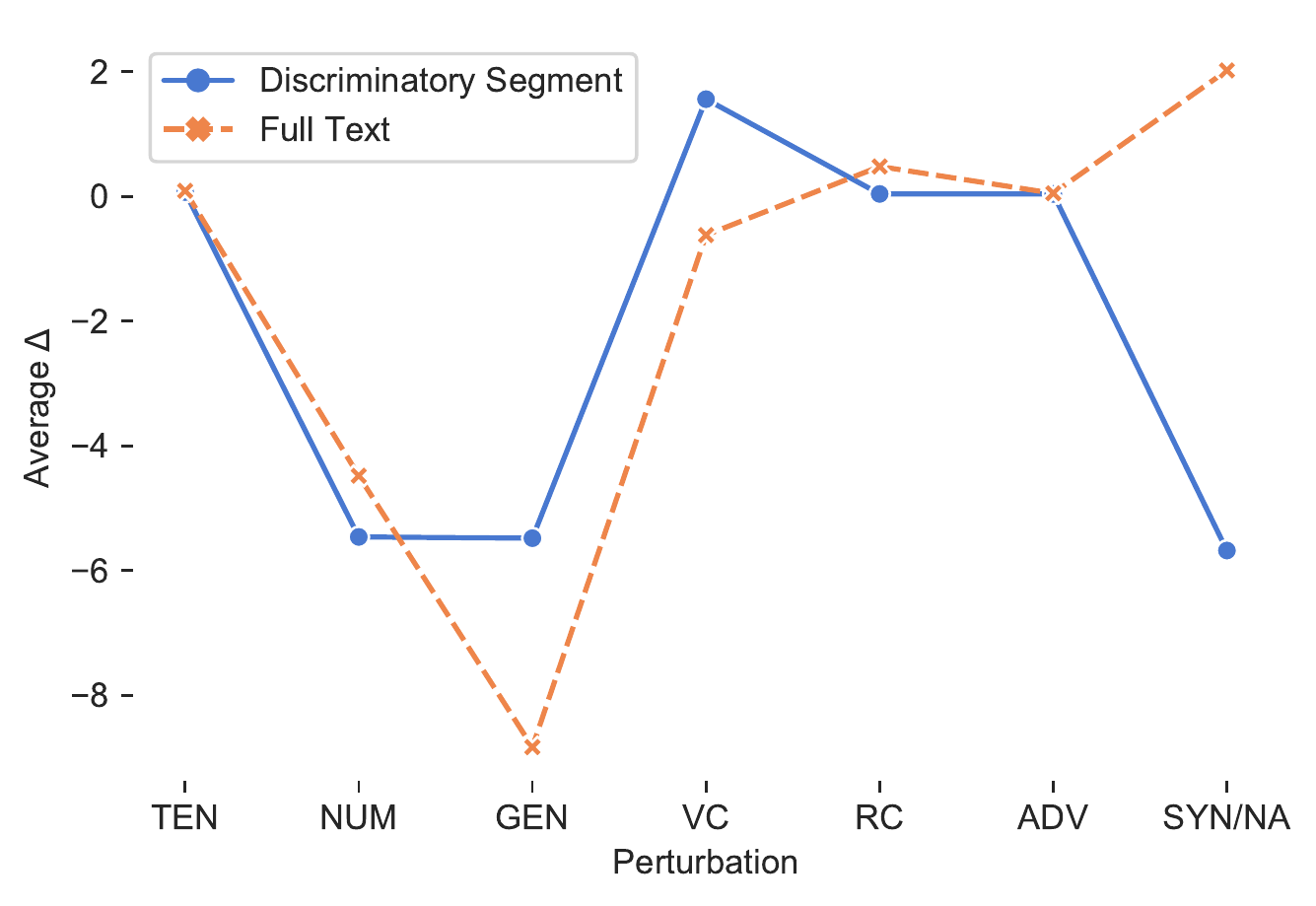}
    \caption{PMI divergence from the original WSC examples in average $\Delta$ for each perturbation. Values below 0 indicate that the difference in PMI between the correct candidate and the incorrect one decreased.}
    \label{fig:pmi_diff}
\end{figure}

\paragraph{Quantifying Associativity}
To verify that the perturbations have not affected the correctness of the original problems with regards to pitfall \textbf{A}, we employ pointwise mutual information (\textsc{PMI}) to test the associativity of both the original and perturbed examples. \textsc{PMI} is known to be a reasonable measure of associativity \cite{church1990word} and, among a variety of measures, has been shown to correlate best with association scores from human judgements of contextual word association \cite{frassinelli2015effect}. We compute unigram \textsc{PMI} on the two corpora used to train \textsc{BERT} (see Appendix \ref{app:pmi} for details). Figure \ref{fig:pmi_diff} shows the \textit{divergence} of the perturbed examples from the original WSC dataset. We estimate divergence as the average difference in \textsc{PMI} between the correct ($\mathcal{C}$) and incorrect ($\mathcal{I}$) candidates: $\Delta = pmi(c_j, x_j) - pmi(i_j, x_j)$ where $\mathcal{X}$ is either: i) the discriminatory segments or ii) the full text of the example, and $pmi(\cdot, \cdot)$ is average unigram PMI. $\Delta$ can be seen as a measure of whether the correct or incorrect candidate is a better `associative fit' for either the discriminatory segment or the full context, making the examples trivial to resolve. Observe that this difference in \texttt{PMI} declines for the perturbed examples, showing that these the perturbed example do not increase in associativity. 

\paragraph{Confirming Solvability}
 Three expert annotators\footnote{Graduate students of linguistics.} are asked to solve the small subset of examples (99 in total across perturbations) which were annotated incorrectly by the majority vote of Mechanical Turk workers. To address pitfall \textbf{B}, the expert annotators are asked to both attempt to solve the instances and indicate if they believe them to be \textit{too ambiguous} to be solved. The majority vote of the annotators determines the preferred referent or whether an instance is ambiguous. Out of a total of 99 examples, 10 were found to be ambiguous. Of the remaining 89 examples, 67 were answered correctly by the majority vote. See Appendix \ref{app:solvability} for more details.
\section{Experimental Protocol}
\label{experiments}

Our experiments are designed to test the robustness of language models to the Winograd Schema perturbations described in the previous section.

\paragraph{Evaluation}
Models are evaluated using two types of measures. The first is  \textit{accuracy}. For each of the perturbations, we report (a) the accuracy on the perturbed set (\textbf{Perturbation accuracy}), (b) the difference in accuracy on the perturbed set and on the \textit{equivalent subset} of original dataset:\footnote{Recall that is was not possible to perturb all examples.} $\Delta_\textbf{Acc.} = \textbf{Perturbation accuracy} - \textbf{Original subset accuracy}$, and (c) \textbf{Pair accuracy}, defined as the number of pairs for which both examples in the pair are correctly answered divided by the total number of pairs. 

The second measure is \textit{stability}, $S$. This is the proportion of perturbed examples $\mathcal{P'}$ for which the predicted referent is the same as the original prediction $\mathcal{P}$: \begin{equation*}\label{eq:stability}
S = \frac{\mid\{(p'_i, p_i) \mid p'_i \in \mathcal{P'} \land p_i \in \mathcal{P} \land p'_i = p_i\}\mid}  {\mid \mathcal{P} \mid}
\end{equation*}
 Since the perturbations do not alter the correct referent, this provides a strong indication of robustness towards them.  

\paragraph{Baseline} 
We take the unigram \texttt{PMI} between candidates and discriminatory segments (see \S\ref{subsec:confounds}) as a baseline. We expect that this simple baseline will perform well for instances with a high level of associativity but not otherwise. 

\paragraph{Language Models}
Our analysis is applied to three out-of-the-box language models (LMs): \textsc{BERT} \cite{devlin2018bert}, \textsc{RoBERTa} \cite{liu2019roberta}, and \textsc{XLNet} \cite{yang2019xlnet}. These models are considered to be the state-of-the-art for the wide variety of natural language understanding tasks found in the GLUE \cite{wang2018glue} and SuperGLUE \cite{superglue} benchmarks. We use the \textit{large} pre-trained publicly available models \cite{Wolf2019HuggingFacesTS}.\footnote{\url{https://github.com/huggingface/pytorch-transformers}}

\paragraph{Fine-tuned Language Models}
We also examine the effect of fine-tuning language models. \textsc{BERT+WW} uses BERT fine-tuned on the MaskedWiki and WscR datasets which consist of 2.4M and 1322 examples \cite{kocijan2019surprisingly}, and RoBERTa+WG is fine-tuned on WinoGrande \textbf{XL}, which consists of 40,938 adversarially filtered examples \cite{sakaguchi2019winogrande}. Both fine-tuned models have been reported by recent work to achieve significant improvements on the WSC. 

\paragraph{Scoring}

To score the two candidate referents in each WSC instance we employ one of two mechanisms. The first, proposed in \newcite{trinh2018simple} and adapted to masked LMs by \newcite{kocijan2019surprisingly} involves computing the probability of the two candidates $c1$ and $c2$, given the rest of the text in the instance $s$. To accomplish this, the pronoun of interest is replaced with a number of \texttt{MASK} tokens corresponding to the number of tokens in each of $c1$ and $c2$. The probability of a candidate, $p(c|s)$ is then computed as the average of the probabilities assigned by the model to the candidate's tokens and the maximum probability candidate is taken as the answer. This scoring method is used for all models, except \textsc{RoBERTa+WG}. For that, we follow the scoring strategy employed in \newcite{sakaguchi2019winogrande} where an instance is split into context and option using the candidate answer as a delimiter.\footnote{\texttt{[CLS] context [SEP]
option [SEP]}, e.g. \textit{[CLS] The sculpture rolled off the shelf because \_\_\_\_ [SEP] wasn't anchored [SEP].} The blank is filled with either option 1 (\textit{the sculpture}) or 2 (\textit{the trophy}).}

\section{Results and Analysis}\label{analysis}

\begin{table*}

\centering
\renewcommand{\arraystretch}{1.1}
\begin{adjustbox}{width=\linewidth}
\begin{tabular}{lcccccccccc}

\toprule
  & \textsc{orig} & \textsc{ten} & \textsc{num} & \textsc{gen} & \textsc{vc} & \textsc{rc} & \textsc{adv} & \textsc{syn/na} &  $Avg$ &  $Avg$ $\Delta_\textbf{Acc.}$ \\
 \cmidrule(lr){2-2} \cmidrule(lr){3-9} \cmidrule(lr){10-10} \cmidrule{11-11}
 \textsc{PMI} & $54.38$ & $54.09$ & $52.96$ & $57.42$ & $54.09$  & $54.41$ & $54.41$ & $51.92$ & $54.24$ & $-2.13$ \\  \midrule
\textsc{Bert} & $61.75$ & $61.92$ & $57.31$ & $57.42$ & $63.64$  &  $62.19$ & $61.48$ & $58.59$ & $60.41$ & $-1.26$ \\

\textsc{XLNet}  & $64.56$ & $60.14$ & $62.45$ & $62.58$ & $57.73$  &  $62.9$ & $64.31$ & $61.05$ & $61.59$ & $-2.78$\\


\textsc{RoBerta}&  $69.82$ & $69.40$ & $64.43$ & $53.55$ & $66.82$  &  $68.55$ & $69.61$ & $57.54$ &  $64.27$ &  $-5.16$ \\
\midrule

\textsc{Bert+WW} & $72.28$ & $70.46$ & $71.15$ & $74.84$ & $65.91$  &  $64.31$ & $72.44$ & $70.88$ & $70.00$ & $-2.82$ \\

\textsc{RoBERTa+WG} & $88.42$ & $89.32$ & $88.53$ & $86.45$ & $83.63$  &  $86.93$ & $88.7$ & $89.05$ & $87.62$ & $-1.06$ \\  \midrule 


\textsc{Humans}& $97.89$ & $96.79$ & $94.46$ & $92.25$ & $92.27$ & $91.16$ & $95.40$ &  $96.14$ & $94.41$ & $-3.83$ \\
 \bottomrule
\end{tabular}
\end{adjustbox}
\caption{Original dataset accuracy (\textsc{orig}) and \textbf{Perturbation accuracy} results for all models and humans. The penultimate column shows the average \textbf{Perturbation accuracy} results.  The rightmost column shows the $\Delta_\textbf{Acc.}$ results, averaged over all perturbations.}

\label{table:accuracy}
\end{table*}

Following the experimental protocol, we evaluate the three out-of-the-box language models and the two fine-tuned models on the original WSC and each of the perturbed sets. Table \ref{table:accuracy} shows \textbf{Perturbation accuracy} results for all models\footnote{It is interesting to note that XLNet is trained on CommonCrawl which indexes an online version of the original WSC found here: \url{https://cs.nyu.edu/faculty/davise/papers/WinogradSchemas/WS.html.}} and contrasts them with human judgements and the \textsc{PMI} baseline. 

\subsection{Language Models}
Humans maintain a much higher performance compared to out-of-the-box LMs across perturbations. The difference in accuracy between the perturbed and original examples, $\Delta_\textbf{Acc.}$, as defined in Section \ref{experiments} is shown in Figure \ref{fig:Δ_Acc}. 
A general trend of decrease can be observed for both models and humans across the perturbations. This decline in accuracy is on average comparable between models and humans --- with a handful of exceptions. Taking the large gap in absolute accuracy into account, this result might be interpreted in two ways. If a comparison is made relative to the upper bound of performance, human performance has suffered from a larger error increase. Alternately, if we compare relative to the lower bound of performance, then the decline in the already low performance of language models is more meaningful, since 'there is not much more to lose'. 

A more transparent view can be gleaned from the stability results shown in Table \ref{table:stability}. Here it can be seen that the three out-of-the-box LMs are \emph{substantially more likely} to switch predictions due to the perturbations than humans. Furthermore, we observe that the LMs are least stable for word-level perturbations like gender (\textsc{gen}), number  (\textsc{num}), and synonym or name replacement  (\textsc{syn/na}), while humans appear to be most affected by sentence-level ones, such as relative clause insertion (\textsc{rc}) and voice perturbation (\textsc{vc}). 

\subsubsection*{Understanding Language Model Performance}
To better understand the biases acquired through pre-training which are pertinent to this task, we consider a) a case of essential feature omission and b) the marginal cases where LMs answer very correctly or incorrectly, in both the original and perturbed datasets. We present analysis for \textsc{BERT}, but similar findings hold for the other LMs.

\paragraph{Masking discriminatory segments}\hspace{-1em} result in identical sentence pairs because these segments are the only part of a sentence that sets WSC pairs apart (see Figure \hyperref[fig:sentences]{1a}). To determine whether there is a bias in the selectional preference for one of the candidates over the other, we test BERT on examples where these discriminatory segments have been replaced with the \texttt{MASK} token. An unbiased model should be close to random selection but \textsc{BERT} consistently prefers (by a margin of $\sim$25-30\%) the candidate which appears second in the text to the one appearing first, for all perturbations except voice, where it prefers the first. This observation holds even when the two referents are inverted, which is possible for the 'switchable' subset of the examples as shown in \citet{trichelair2018evaluation}. This indicates that the selections are not purely semantic but also syntactic or structural and it points towards BERT having a preference referents in the object role. Detailed results are presented in Appendix \ref{app:prefs}.

\begin{table*}[h!]
\centering
\renewcommand{\arraystretch}{1.1}
\begin{tabular}{lcccccccl}
\toprule
  & \textsc{ten} & \textsc{num} & \textsc{gen} & \textsc{vc} & \textsc{rc} & \textsc{adv} & \textsc{syn/na} &  $Avg$ \\
 \midrule
\textsc{PMI}  & $100$ & $100$ & $73.91$ & $100$ & $100$ & $100$ & $100$ & $96.27$ \\  \midrule
\textsc{Bert}  & $89.32$ & $69.17$ & $88.39$ & $79.55$ & $83.75$ & $91.87$ & $68.42$ & $81.40$ \\
\textsc{XLNet}  & $82.21$ & $69.17$ & $66.45$ & $69.55$ & $78.45$ & $84.81$ & $70.53$ & $75.02$\\
\textsc{RoBERTa}  & $91.46$ & $77.47$ & $61.29$ & $79.09$ & $83.75$ & $89.75$ & $68.77$ & $79.26$ \\ \midrule
\textsc{Bert+WW} & $90.04$ & $83.00$ & $89.68$ & $80.45$ & $81.98$ & $92.93$ & $85.96$ & $85.14$ \\ 
\textsc{RoBERTa+WG}  & $96.08$ & $94.07$ & $97.41$ & $91.36$ & $92.22$ & $94.69$ & $96.11$ & $95.24$ \\ \midrule


\textsc{Humans}  & $96.70$ & $94.9$ & $92.9$ & $91.18$ & $91.11$ & $96.11$ & $96.1$ & $94.31$ \\
\bottomrule
\end{tabular}
\caption{Stability results for all models and humans.}
\label{table:stability}
\end{table*}

\paragraph{Marginal examples}\hspace{-1em} are found where the model assigns a much higher probability to one referent over the other. We extract the top 15\% examples where the correct candidate is preferred by the largest margin ($\mathtt{P_{correct} \gg P_{incorrect}}$) and the bottom 15\% where the incorrect one is preferred ($\mathtt{P_{incorrect} \gg P_{correct}}$). Surprisingly, we find that there is a large overlap (50\%--60\%) between these two sets of examples, both in the original and the perturbed datasets.\footnote{To clarify, consider the following original WSC pair: 

\ex. Alice looked for her friend Jade in the crowd. Since \red{she} always \underline{has good luck}, Alice spotted her quickly. 

\ex. Alice looked for her friend Jade in the crowd. Since \red{she} always \underline{wears a red turban}, Alice spotted her quickly. 

The first example gives $\mathtt{P_{correct} \gg P_{incorrect}}$ by the largest margin, and its counterpart gives $\mathtt{P_{incorrect} \gg P_{correct}}$ by the largest margin. In other words, the model assigns a \emph{much higher probability} for \textit{Alice} in both cases.} For the examples which are both the most correct and incorrect, \textsc{BERT} strongly prefers one of the candidates without considering the \underline{special discriminatory segment} which \emph{flips} the correct referent. Indeed we find that the correlation between the probability assigned by BERT to a referent when it is the correct referent and when it is not is very strong and significant, with Spearman's $\rho$ $\approx 0.75$ across perturbations (see Appendix \ref{app:corr_right_wrong} for details). 

\subsection{The effect of fine-tuning}
\label{subsec:ft}
The accuracy and stability results (Tables \ref{table:accuracy} and \ref{table:stability}) indicate that fine-tuning makes language models more robust to the perturbations. \textsc{RoBERTa+WG}, in particular, is the most accurate and most stable model. While impressive, this is not entirely surprising: fine-tuning on task-specific datasets is a well-tested recipe for bias correction \cite{belinkov2019don}. Indeed, these results provide evidence that it is possible to construct larger fine-tuning datasets whose distribution is correct for the WSC. We note that both fine-tuned models perform worst on the \textsc{VC} and \textsc{RC} perturbations, which may not frequently occur in the crowd-sourced datasets used for fine-tuning. To test this intuition, we apply a dependency parser (UDPipe \citep{straka2016udpipe}) to the  WinoGrande XL examples, finding that only $\sim5\%$ of the examples are in the passive voice and  $\sim6.5\%$  contain relative clauses.

\begin{figure}[t]
    \centering
    \includegraphics[scale=0.50]{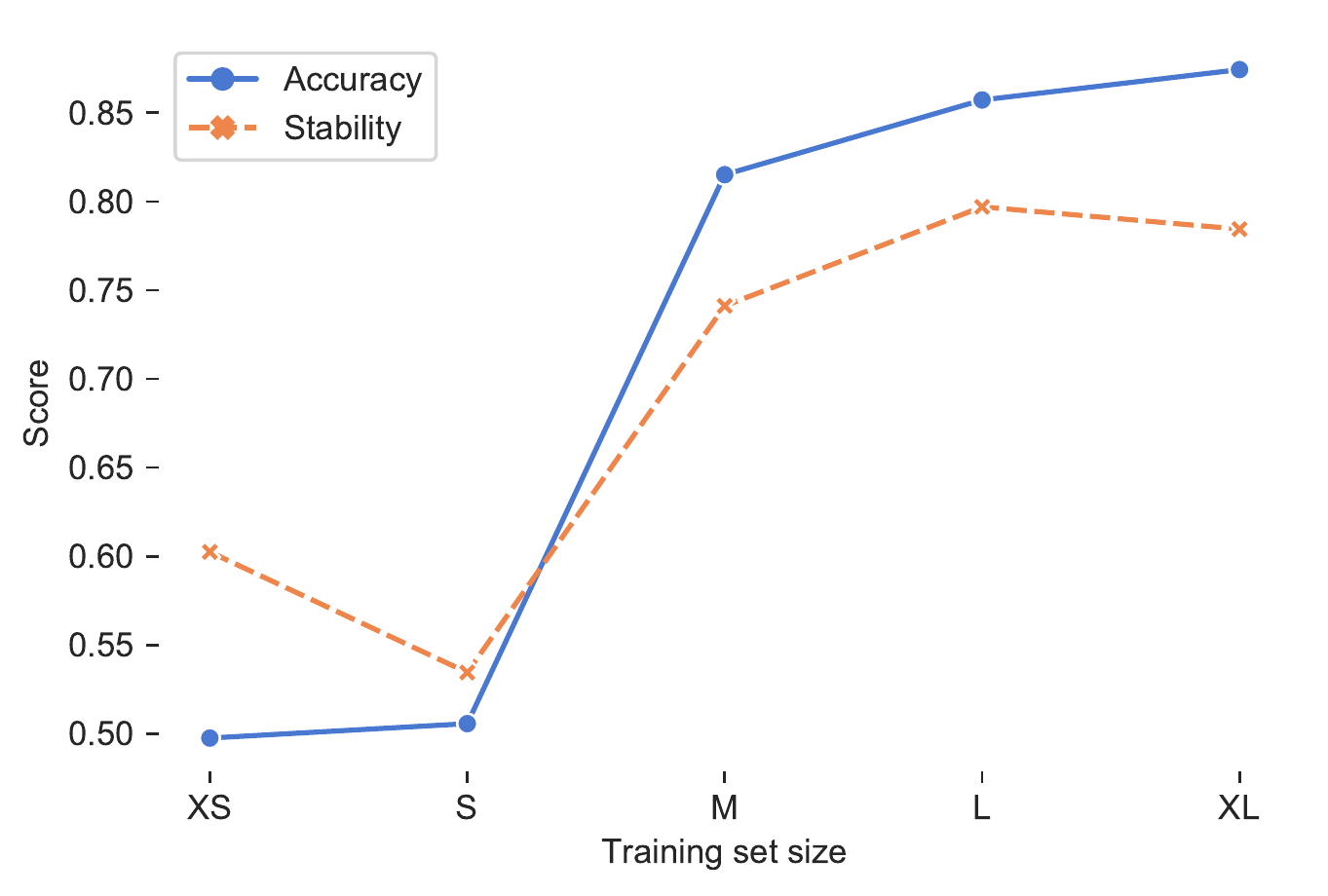}
    \caption{Accuracy and stability scores (averaged across perturbations) for \textsc{RoBERTA} when fine-tuned on five increasing training split sizes.}
    \label{fig:finetuning_analysis}
\end{figure}

\paragraph{How much fine-tuning data is needed?} To quantify the amount of fine-tuning data needed to achieve robustness, we fine-tune \textsc{RoBERTA} on the five WinoGrande training set splits defined by \newcite{sakaguchi2019winogrande}: \textbf{XS} (160)\footnote{No. of examples in set.}, \textbf{S} (640), \textbf{M} (2558), \textbf{L} (10234), and \textbf{XL} (40398). Figure \ref{fig:finetuning_analysis} shows the average accuracy and stability scores for the models fine-tuned on each of the training splits\footnote{Note that the stability score for the model fine-tuned on \textbf{XL} in Figure \ref{fig:finetuning_analysis} is different from that reported in Table \ref{table:stability}. In the latter we reported results from the model provided by \newcite{sakaguchi2019winogrande}, rather than the model we fine-tuned ourselves. Since we utilise identical hyperparameters to theirs for fine-tuning, this anomalous difference in score may perhaps be explained by a difference in initialization as suggested in \newcite{dodge2020fine}.}. We observe that the two smallest splits do not have a sufficient number of examples to adequately bias the classification head, leading to near-random performance. The model fine-tuned on the \textbf{M} split---with just 2558 examples---is, however, already able to vastly outperform the non-fine-tuned \textsc{RoBERTA}. Increasing the number of examples five-fold and twenty-fold leads to significant but fast diminishing improvements.

\begin{figure}[ht]
    \centering
    \includegraphics[scale=0.58]{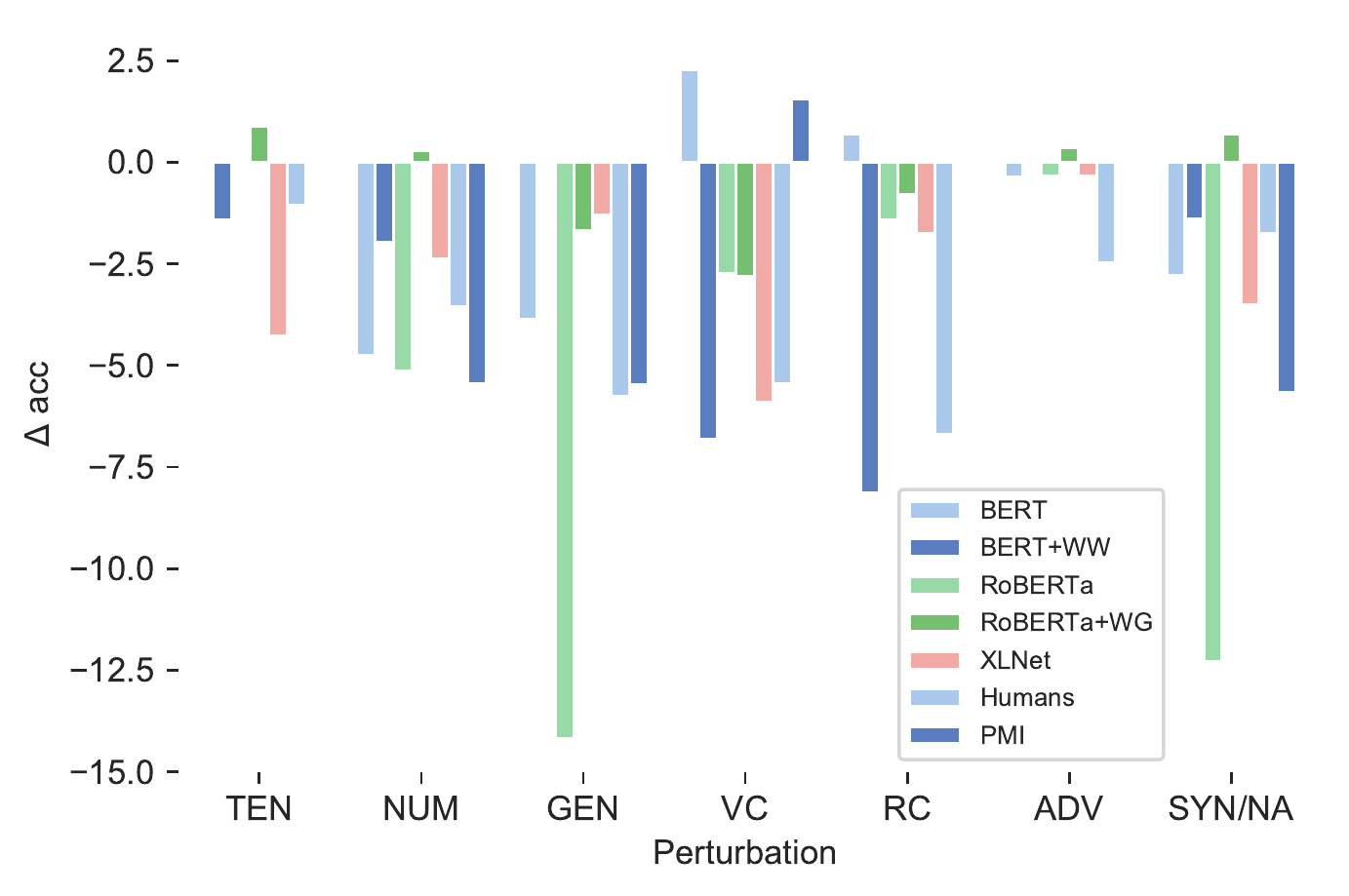}
    \caption{$\Delta_\textbf{Acc.}$ results for all models across perturbations. Values below the x-axis indicate a decline in accuracy compared to the original dataset.}
    \label{fig:Δ_Acc}
\end{figure}

\begin{figure}[t]
    \centering
    \includegraphics[scale=0.58]{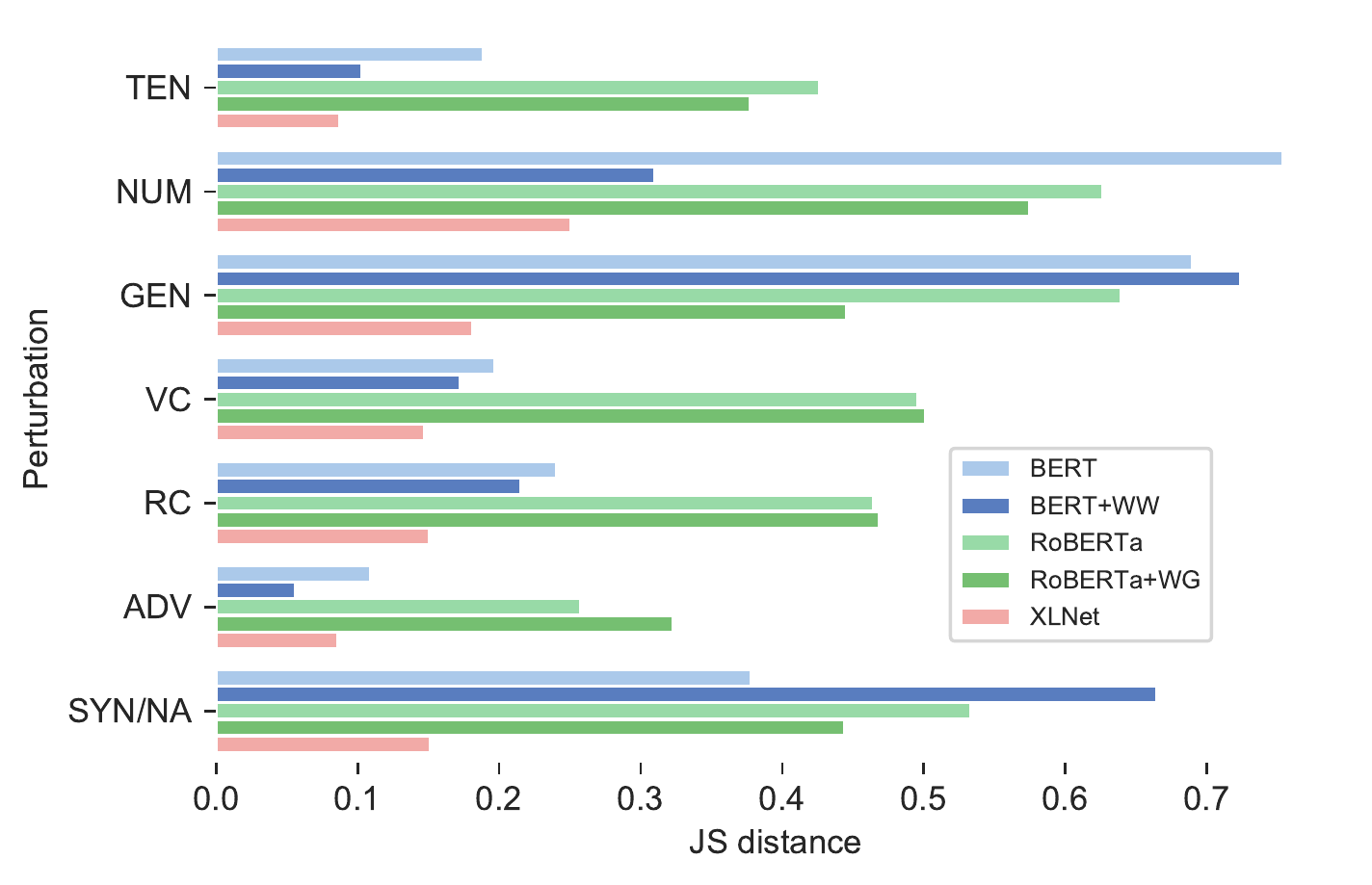}
    \caption{Jensen-Shannon distance between the original and perturbed examples when masking the pronoun of interest.}
    \label{fig:js_dist}
\end{figure}

\paragraph{How do perturbations affect token probability distributions?} To obtain a holistic view of the effect the perturbations have on LMs and fine-tuned LMs, we analyze of the shift in the probability distribution (over the entire vocabulary) which a model assigns to a \texttt{MASK} token inserted in place of the pronoun of interest. We apply probability distribution truncation with a threshold of $p=0.9$ as proposed in \newcite{holtzman2019curious} to filter out the uninformative tail of the distribution. Following this, we compute the Jensen--Shannon distance between this dynamically truncated distribution for an original example and each of its perturbed counterparts. Figure \ref{fig:js_dist} shows the average of this measure over the subset of the 128 examples which are common to all perturbations. Overall, we observe that large shifts in the distribution correspond to lower stability and accuracy scores and that fine-tuned models exhibit lower shifts than their non-fine-tuned counterparts. The difference in shifts between out-of-the-box models and their fine-tuned counterparts is lower for the VC, RC and ADV perturbations, meaning that when fine-tuned, the models' probability distributions are roughly just as divergent for these perturbations as they were before fine-tuning. We hypothesize the same reasons we did in~\ref{subsec:ft}, which is that these examples are just under-represented in our fine-tuning corpus; indeed, these results roughly correspond to the differences in $\Delta_\textbf{Acc.}$ from Figure~\ref{fig:Δ_Acc}. 

Further details about the number of examples excluded via the probability distribution truncation and other measures of the perturbations' effect can be found in Appendix \ref{app:changes}. 

\subsection{Error Analysis}
\paragraph{Pair Accuracy} Here we consider a more challenging evaluation setting where each WSC pair is treated as a single instance. Since the WSC examples are constructed as minimally contrastive pairs \cite{levesque2012winograd}, we argue that this is an appropriate standard of evaluation. Consider again the example in Figure \hyperref[fig:sentences]{1a}. It is reasonable to suppose that for an answerer which truly `understands' \cite{levesque2012winograd}, being able to link the concepts \textit{heavy} and \textit{son} in one of the resolutions is closely related and complementary to linking the concepts \textit{weak} and \textit{man} in the other.\footnote{As a sanity check, consider random pairings of WSC examples. There is no such complement.} 

\begin{figure}[t]
    \centering
    \includegraphics[scale=0.45]{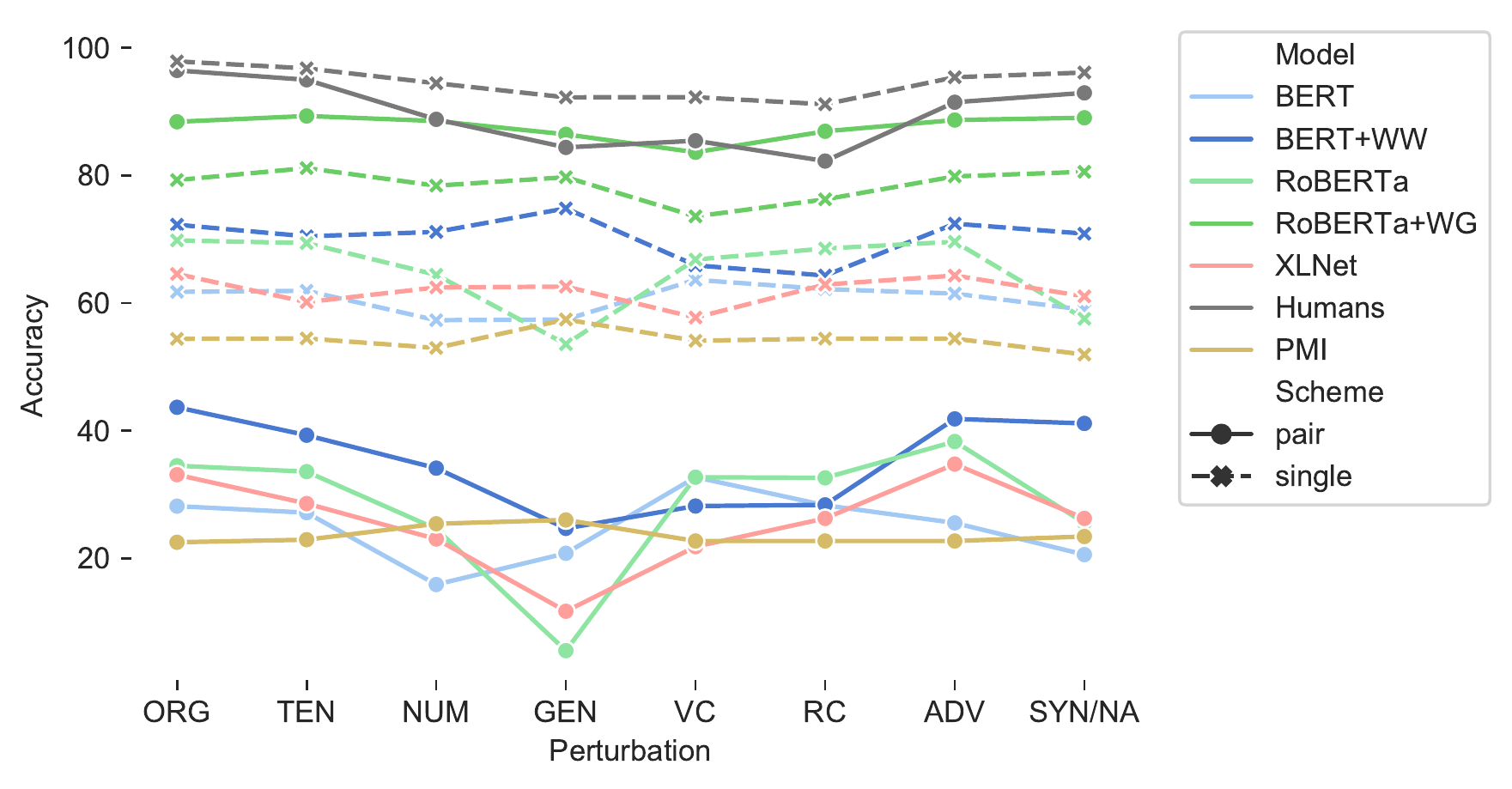}
    \caption{\textbf{Pair accuracy} and \textbf{Perturbation accuracy} results. The latter are labeled as \textit{single}.}
    \label{fig:pair_acc}
\end{figure}

The results for this evaluation are shown in Figure \ref{fig:pair_acc}. They show that human resolution of the problems exhibits greater complementarity compared to the language models; human pair accuracy (pair) is closer to perturbation accuracy (single) than is the case for the LMs. Furthermore, human performance on pair accuracy is more robust to perturbations when compared to the models. Indeed, the large gap between pair accuracy and perturbation accuracy raises some doubts about the performance of these models. 
However, \textsc{RoBERTa-WG} is a notable exception, showing near-human robustness to pair complementarity.


\begin{figure}[ht]
    \centering
    \includegraphics[scale=0.46]{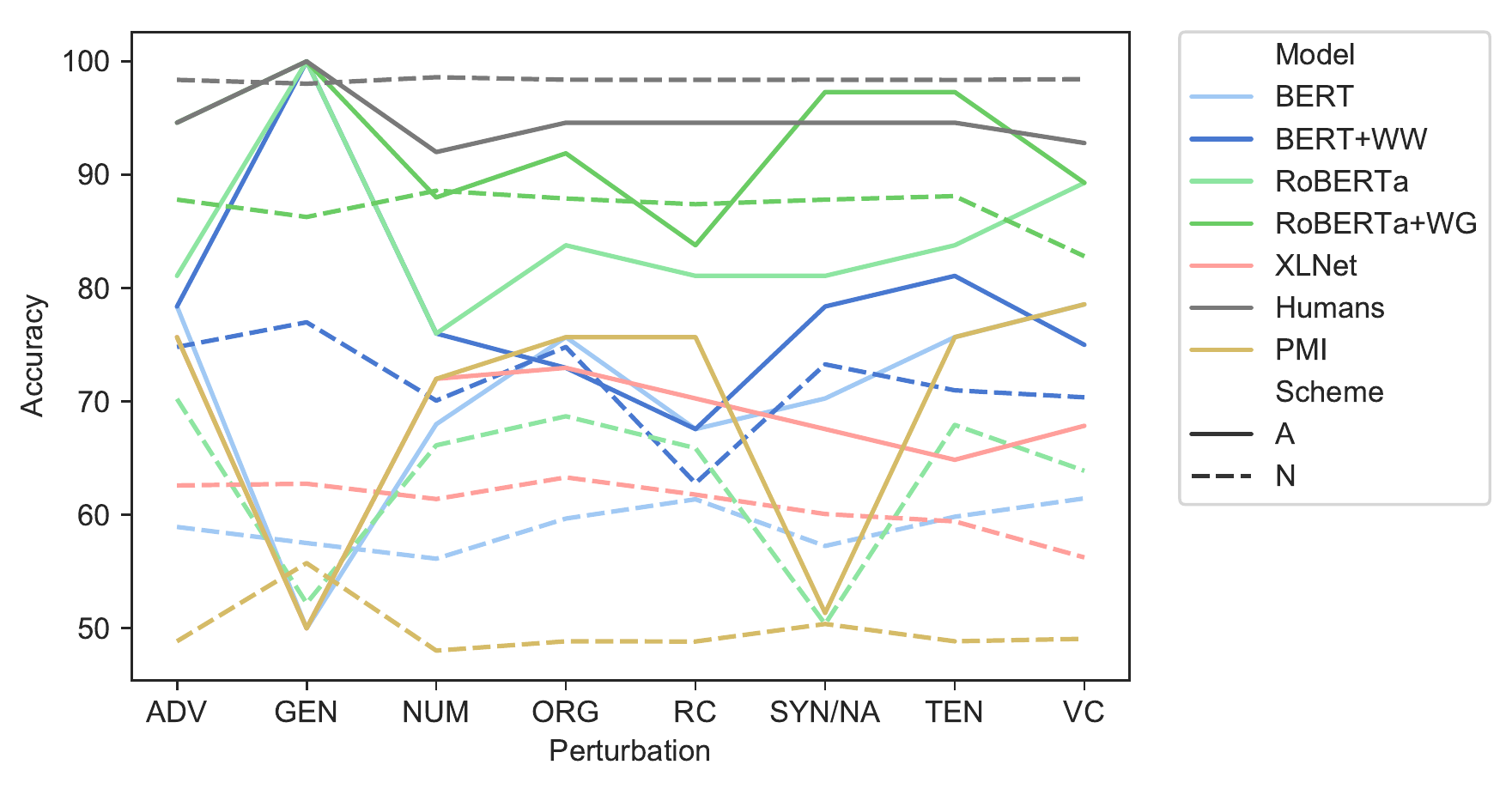}
    \caption{Perturbation accuracy on the Associative (A) and Non-Associative (N) subsets of the data.}
    \label{fig:error_analysis_assoc_acc}
\end{figure}

\paragraph{Associativity}
Next, we examine the effect of associativity on performance. Figure \ref{fig:error_analysis_assoc_acc} shows accuracy results\footnote{Note that the large variance in results on the associative subset of gender is due to it consisting of only two examples.} for all perturbations on the associative and non-associative subsets of the WSC as labelled by \newcite{trichelair2018evaluation}. We observe that the difference between associative and non-associative is much smaller for humans and that unlike all language models, humans do better on the former than the latter. As expected, the \textsc{PMI} baseline does almost as well as the LMs on the associative subset but it performs at chance level for the non-associative subset.

\section{Conclusion}
\label{conclusion}

We presented a detailed investigation of the effect of linguistic perturbations on how language models and humans perform on the Winograd Schema Challenge. We found that compared to out-of-the-box models, humans are significantly more stable to the perturbations and that they answer non-associative examples with higher accuracy than associative ones, show sensitivity to WSC pair complementarity, and are more sensitive to sentence-level (as opposed to word-level) perturbations. In an analysis of the behaviour of language models, we observe that there is a preference for referents in the object role and that the models do not always consider the \underline{discriminatory segments} of examples. Finally, we find that fine-tuning language models can lead to much-improved accuracy and stability. It remains an open question whether this task-specific approach to generalisation constitutes a true advancement in ``reasoning''.  Fine-tuning a model on a rather large number of examples similar to the WSC leads to increased robustness, but this stands in stark contrast to humans, who are robust to the perturbations without having been exposed to similar examples in the past.

\section*{Acknowledgments}
We would like to thank Mitja Nikolaus, Artur Kulmizev, Ana Valeria Gonzalez, and the anonymous reviewers for their helpful comments. Mostafa Abdou and Anders S{\o}gaard  are supported by a Google Focused Research Award and a Facebook Research Award. Yonatan Belinkov was supported by the Harvard Mind, Brain, and Behavior Initiative.

\bibliography{acl2019}
\bibliographystyle{acl_natbib}
\newpage
\appendix
\section{Observations on original dataset}
\label{app:obs}

\begin{enumerate}
    \item A few of the original examples were of unorthodox design: for instance, consider the pair:
    \eenumsentence{\item Look! There is a minnow swimming right below that duck! It had better get away to safety fast!
    \item Look! There is a shark swimming right below that duck! It had better get away to safety fast!}
    
    Here, instead of having a discriminatory segment select which of the two nouns could be the antecedent, one of the nouns is switched out with another.
    \item Example 90 has a typo in the question where Kamchatka is spelled as `Kamtchatka'.
\end{enumerate}
\section{Human Judgements}
\label{app:humanresults}
\begin{figure*}[t]
    \centering
    \includegraphics[width=0.99\textwidth, height=0.6\textheight]{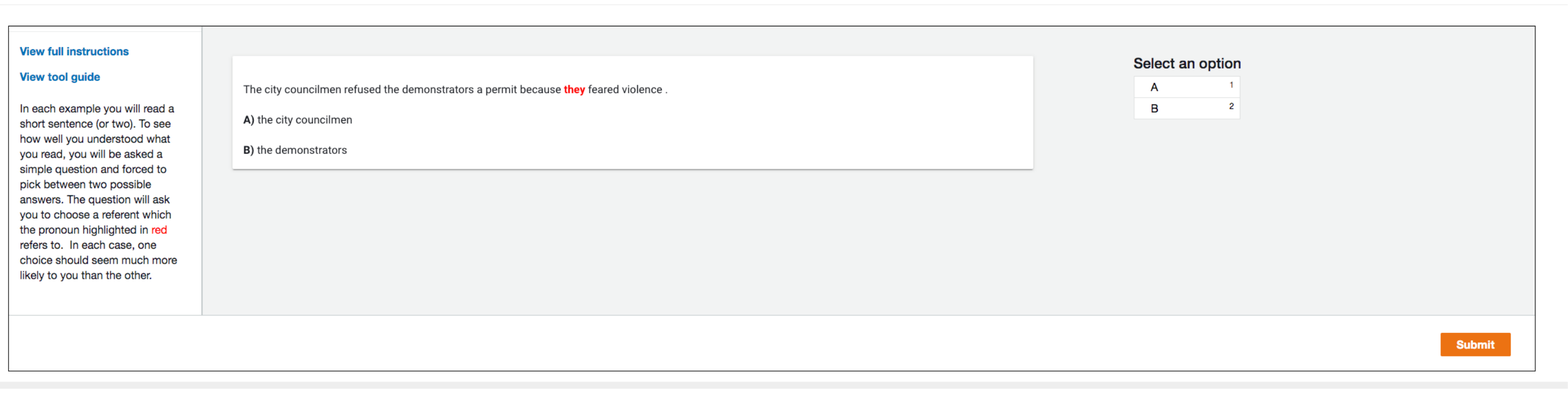}
        \vspace{-45mm}
    \caption{Sample of Mturk template shown to annotators.}
    \label{fig:interface_mturk}
\end{figure*}

Table \ref{tab:humanresults} shows the proportion of instances for which all three annotators agreed and the average time required by annotators for the original examples and each of the perturbed datasets. Figure \ref{fig:interface_mturk} shows the Amazon Mechanical Turk template used. The annotator pool was restricted to native speakers of English located in the United States who were classified by Mturk as `masters' and had a HITs approval rate above 99\%.
\begin{table}[t]

\centering
\begin{tabular}{@{}lcc@{}} \toprule
\bf Pert. & \bf Full Agreement & \bf Avg. Time \\ \midrule
\textsc{org} & 82.45 & 15.32  \\
\hline
\textsc{ten} &  82.91 & 16.39 \\ 
\textsc{num} & 83.00 & 19.56 \\ 
\textsc{gen} & 78.06 & 19.24 \\
\textsc{vc}  & 82.72 & 17.02 \\
\textsc{rc}  & 82.68 & 17.83 \\
\textsc{adv} & 82.68 & 17.69 \\
\textsc{syn/na} & 82.45 & 15.26 \\ 
\midrule
\bottomrule
\end{tabular}
\caption{Annotation statistics: Proportion of examples with full agreement and average time required for answering in seconds.}
\label{tab:humanresults}
\end{table}

\section{Pointwise Mutual Information}
\label{app:pmi}
We compute unigram Pointwise Mutual Information statistics using the Hyperwords\footnote{https://bitbucket.org/omerlevy/hyperwords/} package \cite{levy2015improving}. If a corpus is split into a collection $D$ of words $W$ and their contexts $C$, we can compute co-occurrence counts for each pair of $w \in W$ and $c \in C$. PMI is then defined as the log-ratio between the joint
probability of $w$ with $c$ and the product of their marginal probabilities. Refer to \newcite{levy2015improving} for further details. For generating a collection $D$ of word-context pairs, we use the following hyperparameter settings: a minimal word count of 200 for being in the vocabulary, a context window size of 6, dynamic context windows, positional contexts (where each context is a conjunction of a word and its relative position to the target
word).

\section{Confirming Solvability}
\label{app:solvability}

Table \ref{tab:confirming_solv} shows the breakdown by perturbation type of the expert annotations which were gathered for examples that were annotated incorrectly by the Mechanical Turk workers. 
\begin{table}
\centering
\begin{tabular}{@{}lcccc@{}} \toprule
Counts & \bf All & \bf Ambig. & \bf Non-Ambig. & \bf Correct\\ \midrule
\textsc{ten} &  9 & 0 & 9  & 8 \\ 
\textsc{num} & 14 & 2 & 12  & 9\\ 
\textsc{gen} & 12 & 2 & 10  & 10\\
\textsc{vc}  & 17 & 3 & 14  & 12\\
\textsc{rc}  & 25 & 1 & 24  & 13\\
\textsc{adv} & 13 & 0 & 13  & 11\\
\textsc{syn/na} & 9 & 2 & 7 & 4  \\ 
\midrule
\bottomrule
\end{tabular}
\caption{Breakdown of solvability annotation counts by perturbation. \textbf{Ambig.} indicates the count of examples labeled as Ambiguous, \textbf{Non-Ambig.} is the number of remaining examples. \textbf{Correct} indicates the number of those which is solved correctly. }
\label{tab:confirming_solv}
\end{table}
\section{Notes on construction of perturbed dataset}
\label{app:notes_perturbations}
\paragraph{Tense switch (\textsc{ten})} 
Examples 168--172 could not be changed while maintaining the semantics of the instance intact.

\paragraph{Relative clause insertion (\textsc{rc})}
The pre-selected set of 19 templates is shown below:
\begin{itemize}
    \item ``who we had discussed \_\_"
    \item ``who he had discussed \_\_" 
    \item ``who she had discussed \_\_" 
    \item ``who you had discussed \_\_"
    \item ``which we had seen \_\_" 
    \item ``which he had seen \_\_" 
    \item ``which she had seen \_\_" 
    \item ``which you had seen \_\_"
    \item ``who we know from \_\_" 
    \item ``who he knows from \_\_" 
    \item ``who she knows from \_\_" 
    \item ``who you know from \_\_"
    \item ``that is mentioned in \_\_"
    \item ``that is located at \_\_" 
    \item ``that is close to \_\_" 
    \item ``that is known for \_\_" 
    \item ``which had been \_\_",
    \item ``who you met \_\_" 
    \item ``that is \_\_" 
    \item ``which was put there \_\_"
\end{itemize}

\paragraph{Synonym/Name substitution (\textsc{syn/na})}
No appropriate synonyms were found for \textit{tide} and \textit{wind} in examples 130 and 131.

\paragraph{Adverbial qualification (\textsc{adv})}
Two instances (95 and 96) in which the main verb was already modified were excluded. 

\section{Referent preferences}
\label{app:prefs}

Table \ref{table:prefs} shows the percentage of examples in the switchable subset of the datasets where the second referent in the text was assigned a higher probability than the first, for both the original and reversed referent order.

\begin{table}[ht]
    \centering
    \begin{tabular}{lcccc}
        \toprule
        \textbf{Pert.} & \textbf{Original} & \textbf{Reversed}\\
        \midrule
        \textsc{org} & 66.90 & 70.42 \\
        \midrule
        \textsc{ten} & 62.38 & 65.14 \\
        \textsc{num} & 60.16 & 56.10 \\
        \textsc{gen} & 72.17 & 75.65 \\
        \textsc{vc} & 38.14 & 39.83 \\
        \textsc{rc} & 63.57 & 68.57 \\
        \textsc{adv} & 68.08 & 70.92 \\
        \textsc{syn/na} & 59.12 & 64.23 \\
        \bottomrule
    \end{tabular}
    \caption{Percentage of examples in switchable subset with probabilities assigned to the second referent in the text rather than the first, for both the original and reversed referent order.}
    \label{table:prefs}
\end{table}{}
\section{Effect of perturbations}
\label{app:changes}

\paragraph{Nucleus Sampling} Table \ref{tab:vocab_filtrered} shows the average number of vocabulary items kept after Nucleus sampling with
$p=0.9$ is applied.

\begin{figure}[t]
    \centering
    \includegraphics[scale=0.29]{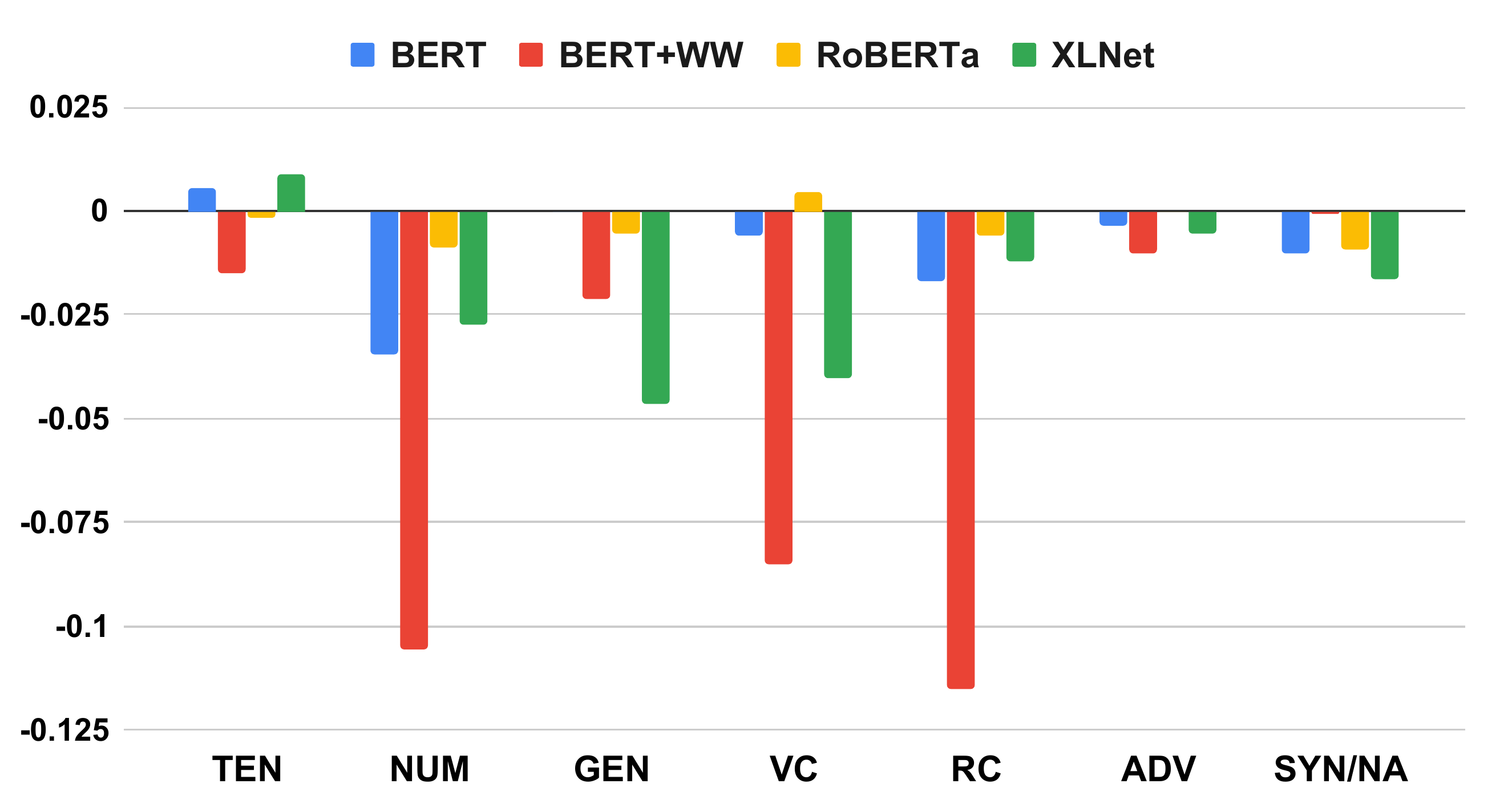}
    \caption{The difference between average probability shift for the correct and the incorrect referents per perturbation. Y-axis values above zero mean the correct referent became more likely on average after a perturbation and vice versa.}
    \label{fig:prob_shift_diff}
\end{figure}

\paragraph{Probability shift} is defined as the difference in the probability of a candidate before and after a perturbation is applied.
Figure \ref{fig:prob_shift_diff} shows the difference in average probability shift between the correct candidates and the incorrect candidates for each of the models per perturbation type. This provides a view that is meaningfully different from accuracy, as the probability of a candidate can shift without exceeding the threshold required to change a model's prediction. We find that there is a general trend of the incorrect candidates becoming more likely relative to the correct ones. This can be seen as confirming that, on average, nearly all perturbations make the problems more difficult for all models.  

\begin{figure}[t]
    \centering
    \includegraphics[scale=0.34]{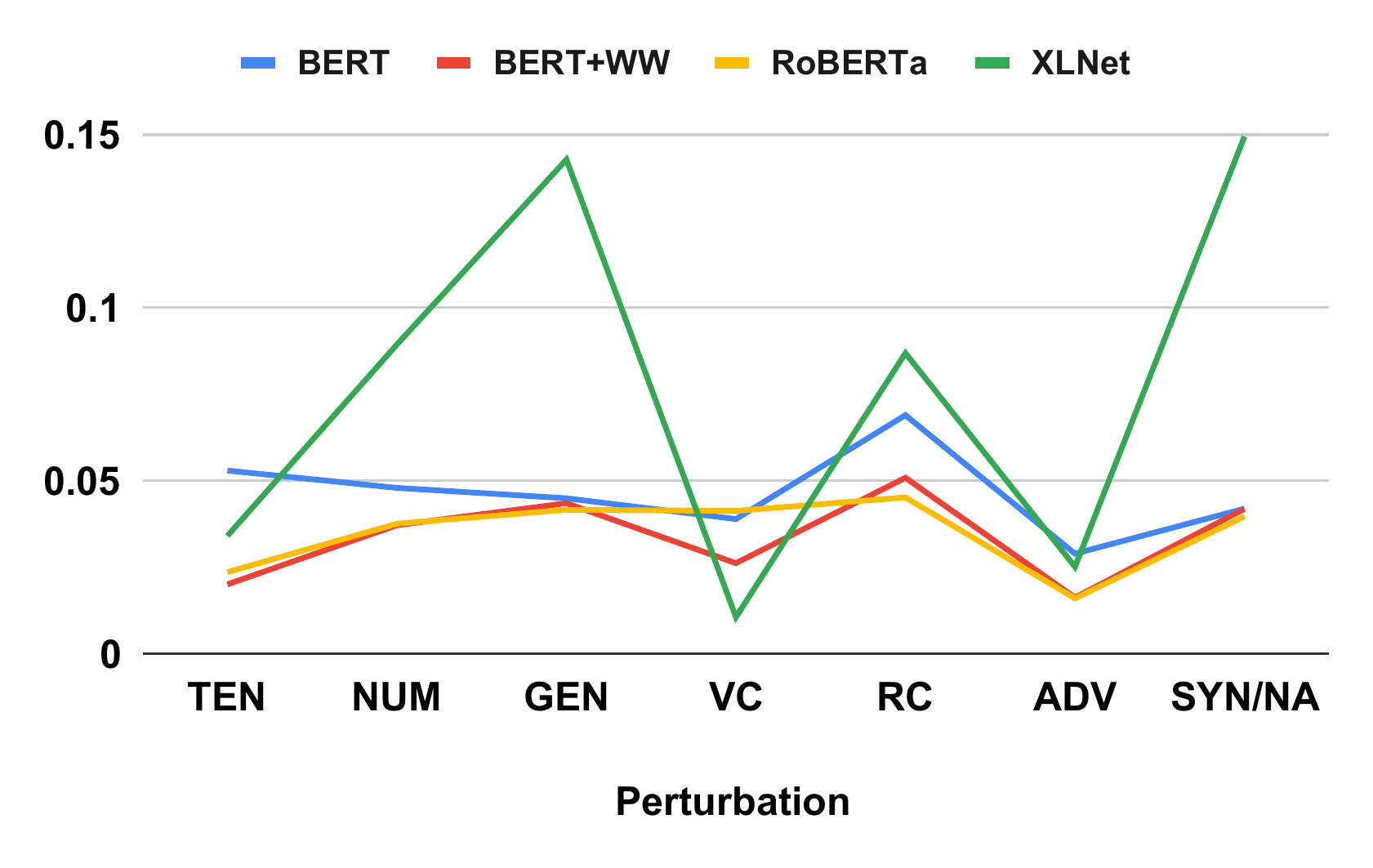}
    \caption{The correlation of pronoun hidden state representation distance from the original for each perturbation.}
    \label{fig:distance_analysis}
\end{figure}

\paragraph{Hidden state representation distance} is used to provide a more holistic view of the correspondence between the representations derived for the different perturbations. The analysis is conducted on the 128 examples which are common between all datasets. A representation is derived for each example by taking the max-pool of hidden-state representations of a model's final layer. For each of the seven perturbations $p$, we compute pairwise correlation distance\footnote{This is preferable to other distance measures as it normalizes both the mean and variance of activity patterns over experimental conditions.} between each pair of original and perturbed example representations yielding a vector $\vec{D_p} \in \mathbb{R}^{128}$. The mean of $\vec{D_p}$ is then computed as an aggregate measure of the distance between the representations derived from a perturbation $p$ and the original $o$. Figure \ref{fig:distance_analysis} shows a plot of this for all perturbations for each of the models.

\begin{table*}
\centering
\begin{tabular}{@{}lccccc@{}} \toprule
 Perturbation & \textsc{BERT} & \textsc{RoBERTa} & \textsc{XLNet} & \textsc{BERT+WW} & \textsc{RoBERTa+WG}\\ \midrule

\hline

\textsc{org}& 19.81 & 203 & 1.26 &	1.07 &	1021.44 \\ 
\textsc{ten}& 23.88	& 165.84 & 1.26 & 1.09 & 947.53 \\ 
\textsc{num} &90.35 & 341.05 &	1.57 &	1.30 &	1087.78\\ 
\textsc{gen} &18.11	& 128.37	& 1.44	& 1.19 &	1039.84\\
\textsc{vc} & 41.88 & 154.21	& 1.28 &	1.09 & 961.04\\
\textsc{rc} & 21.02 &	97.35	& 1.35 &	1.14 &	952.09\\
\textsc{adv} & 17.01 & 145.35 & 1.23 &	1.10 &	1004.14\\
\textsc{syn/na} & 31.50 & 199.26 & 1.39 &	1.11 &	1055.71 \\ 
\midrule
\midrule
\textsc{vocab. size} & 30522& 	50265	& 32000  & 	30522 & 	50265\\ 
\midrule
\bottomrule
\end{tabular}
\caption{Average number of vocabulary items left after probability distribution truncation with $p=0.9$ is applied.}
\label{tab:vocab_filtrered}
\end{table*}
\label{app:corr_right_wrong}
\section{Candidate probability correlations}

\begin{figure}
    \centering
    \includegraphics[width=0.48\textwidth]{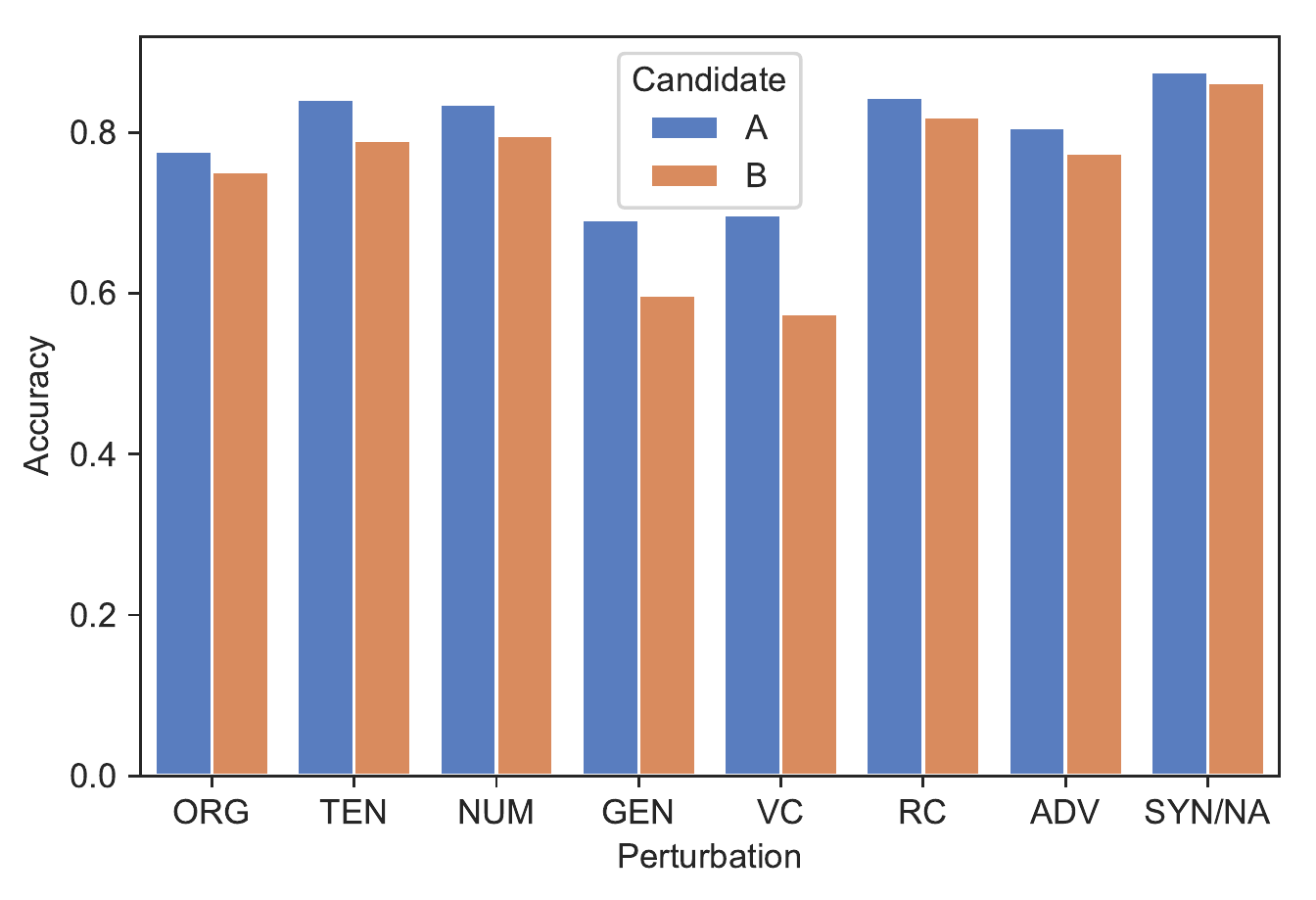}
    \caption{Correlation (Spearman's $\rho$) between the probability of a candidate when it is the correct candidate and when it is the incorrect one. Candidates A and B are the first and second candidates in a WSC instance. }
    \label{fig:corr_right_wrong}
\end{figure}

Figure \ref{fig:corr_right_wrong} shows the average correlation between a candidate's probability when it is the correct referent and when it is not. 

\end{document}